%% file: paper.tex
\documentclass{www2010-accepted}

\usepackage{latexsym,amssymb,amsmath,graphicx,float}
\usepackage{subfigure}
\usepackage{algorithm}
\usepackage{algorithmic}
\usepackage{multirow}
\usepackage{paralist}
\usepackage{times}

\input{math.tex}
\newcommand{\algref}[1]{Algorithm~\ref{#1}}
\newcommand{\figref}[1]{Figure~\ref{#1}}
\newcommand{\tblref}[1]{Table~\ref{#1}}
\newcommand{\secref}[1]{Section~\ref{#1}}
\newcommand{\thmref}[1]{Theorem~\ref{#1}}
\newcommand{\eqnref}[1]{Eq.~(\ref{#1})}
\newcommand{\Aset}{\mathcal{A}}
\newcommand{\Aalg}{\mathsf{A}}
\newcommand{\aka}{{a.k.a.}}
\newcommand{\eg}{\textit{e.g.}}
\newcommand{\etc}{{etc.}}
\newcommand{\ie}{\textit{i.e.}}
\newcommand{\cf}{\textit{c.f.}}
\newcommand{\algfont}[1]{\textsf{#1}}
\newcommand{\subsubsubsection}[1]{\noindent\paragraph{#1}}

\newcommand{\figsqueeze}{\vspace{-0.14in}}

\long\def\comment#1{}

\begin{document}
\title{A Contextual-Bandit Approach to \\ Personalized News Article Recommendation%
}

\numberofauthors{3}

\author{
\alignauthor Lihong Li$^\dag$, Wei Chu$^\dag$,\\
       \affaddr{$^\dag$Yahoo! Labs}\\
       \email{lihong,chuwei@yahoo-inc.com}\\
\alignauthor John Langford$^\ddag$ \\
       \affaddr{$^\ddag$Yahoo! Labs}\\
       \email{jl@yahoo-inc.com}\\
\alignauthor Robert E. Schapire$^+$\titlenote{This work was done while R.~Schapire visited Yahoo! Labs.}\\
       \affaddr{$^+$Dept of Computer Science}\\
       \affaddr{Princeton University}\\
       \email{schapire@cs.princeton.edu}
}
\date{\today} %

\maketitle
\begin{abstract}
Personalized web services strive to adapt their services (advertisements,
news articles, \etc) to individual users by making use of both content
and user information.  Despite a few recent advances, this problem
remains challenging for at least two reasons.  First, web service is
featured with dynamically changing pools of content, rendering
traditional collaborative filtering methods inapplicable.  Second,
the scale of most web services of practical interest calls for solutions
that are both fast in learning and computation.  %

In this work, we model personalized recommendation of news articles
as a contextual bandit problem, a principled approach in which a
learning algorithm sequentially selects articles to serve users based on
contextual information about the users and articles, while simultaneously
adapting its article-selection strategy based on user-click feedback to
maximize total user clicks.

The contributions of this work are three-fold.  First, we propose a
new, general contextual bandit algorithm that is computationally
efficient and well motivated from learning theory. Second, we argue
that any bandit algorithm can be reliably evaluated \emph{offline}
using previously recorded random traffic. Finally, using this offline
evaluation method, we successfully applied our new algorithm to a
Yahoo!\ Front Page Today Module dataset containing over $33$ million events.
Results showed a $12.5\%$ click lift compared to a standard context-free
bandit algorithm, and the advantage becomes even greater when data gets more scarce.
\end{abstract}



\section{Introduction}

This paper addresses the challenge of identifying the most appropriate
web-based content at the best time for individual users.  Most service
vendors acquire and maintain a large amount of content in their
repository, for instance, for filtering news
articles~\cite{Das2007GoogleNews} or for the display of
advertisements~\cite{Anagnostopoulos07Just}.  Moreover, the content	
of such a web-service repository changes dynamically, undergoing
frequent insertions and deletions. In such a setting, it is crucial to
quickly identify interesting content for users.  For instance, a news
filter must promptly identify the popularity of breaking news, while
also adapting to the fading value of existing, aging news stories.

It is generally difficult to model popularity and temporal changes
based solely on content information. In practice, we usually explore
the unknown by collecting consumers' feedback in real time to
evaluate the popularity of new content while monitoring changes in its
value~\cite{Agarwal09Online}. For instance, a small amount of traffic
can be designated for such exploration.  Based on the users' response
(such as clicks) to randomly selected content on this small slice of
traffic, the most popular content can be identified and exploited on
the remaining traffic.  This strategy, with random exploration on an
$\epsilon$ fraction of the traffic and greedy exploitation on the
rest, is known as \algfont{$\epsilon$-greedy}.  Advanced exploration approaches
such as \algfont{EXP3}~\cite{Auer02Nonstochastic} or \algfont{UCB1}~\cite{Auer02Finite}
could be applied as well. Intuitively, we need to distribute more
traffic to new content to learn its value more quickly, and
fewer users to track temporal changes of existing content.

Recently, personalized recommendation has become a desirable feature
for websites to improve user
satisfaction by tailoring content
presentation to suit individual users' needs~\cite{Brusilovsky07Adaptive}.  Personalization
involves a process of gathering and storing user attributes, managing
content assets, and, based on an analysis of current and past
users' %
behavior, delivering the individually best
content to the present user being served.

Often, both users and content are represented by sets of
features. User features may include historical activities at an
aggregated level as well as declared demographic information. Content
features may contain descriptive information and categories. In this
scenario, exploration and exploitation have to be deployed at an
individual level since the views of different users on the same
content can vary significantly.  Since there may be a very large
number of possible choices or actions available, it becomes critical
to recognize commonalities between content items and to transfer that
knowledge across the content pool.

Traditional recommender systems, including collaborative
filtering, content-based filtering and hybrid approaches, can
provide meaningful recommendations at an individual level by
leveraging users' interests as demonstrated by their past
activity. Collaborative filtering \cite{Schafer99Recommender}, by
recognizing similarities across users based on their consumption
history, provides a good recommendation solution to the scenarios
where overlap in historical consumption across users is relatively
high and the content universe is almost static.  Content-based
filtering helps to identify new items which well match an existing
user's consumption profile, but the recommended items are always
similar to the items previously taken by the user
\cite{Mladenic99Textlearning}. Hybrid approaches
\cite{Burke05Hybrid} have been developed by combining two or more
recommendation techniques; for example, the inability of
collaborative filtering to recommend new items is commonly
alleviated by combining it with content-based filtering.

However,
as noted above, in many web-based scenarios, the content universe
undergoes frequent changes, with content popularity changing over
time as well.  Furthermore, a significant number of visitors are
likely to be entirely new with no historical consumption record
whatsoever; this is known as a \emph{cold-start}
situation~\cite{Park2006Naive}.  These
issues make traditional recommender-system approaches difficult to apply,
as shown by prior empirical studies~\cite{Chu09Personalized}.
It thus becomes indispensable to learn the goodness of match
between user interests and content when one or both of them are new.
However, acquiring such information can be expensive and may reduce
user satisfaction in the short term, raising the question of optimally
balancing the two competing goals: maximizing user satisfaction in the long run,
and gathering information about goodness of match between user interests and content.

The above problem is indeed known as a feature-based exploration/exploitation problem.
In this paper, we formulate it as a \emph{contextual bandit} problem,
a principled approach in which a learning algorithm sequentially selects
articles to serve users based on contextual information of the user and
articles, while simultaneously adapting its article-selection strategy
based on user-click feedback to maximize total user clicks in the long run.
We define a bandit problem and then review
some existing approaches in \secref{sec:related}.
Then, we propose a new algorithm, \algfont{LinUCB}, in \secref{sec:linucb}
which has a similar regret analysis to the best known algorithms for
competing with the best linear predictor, with a lower computational
overhead.  We also address the problem of \emph{offline} evaluation in
\secref{sec:evaluation}, showing this is possible for \emph{any}
explore/exploit strategy when interactions are independent and
identically distributed (i.i.d.), as might be a
reasonable assumption for different users.  We then test our
new algorithm and several existing algorithms using this offline
evaluation strategy in \secref{sec:experiments}.

\section{Formulation \& Related Work} \label{sec:related}

In this section, we define the $K$-armed contextual bandit problem formally, and as an example, show how it can model the personalized news article recommendation problem.  We then discuss existing methods and their limitations.

\subsection{A Multi-armed Bandit Formulation} \label{sec:bandit}

The problem of personalized news article recommendation can be naturally modeled as a multi-armed bandit problem with context information.  Following previous work~\cite{Langford08Epoch}, we call it a \emph{contextual bandit}.\footnote{In the
literature, contextual bandits are sometimes called bandits with covariate,
bandits with side information, associative bandits, and associative reinforcement learning.}
Formally, a contextual-bandit algorithm $\Aalg$ proceeds in discrete trials $t=1,2,3,\ldots$  In trial $t$:
\begin{compactenum}
\item{The algorithm observes the current user $u_t$ and a set
$\Aset_t$ of arms or actions together with their feature vectors $\vecb{x}_{t,a}$ for $a\in\Aset_t$.  The vector $\vecb{x}_{t,a}$ summarizes information of \emph{both} the user $u_t$ and arm $a$, and will be referred to as the \emph{context}.}
\item{Based on observed payoffs in previous trials, $\Aalg$ chooses an arm $a_t\in\Aset_t$, and receives payoff $r_{t,a_t}$ whose expectation depends on both the user $u_t$ and the arm $a_t$.}
\item{The algorithm then improves its arm-selection strategy with the new observation, $(\vecb{x}_{t,a_t},a_t,r_{t,a_t})$.  It is important to emphasize here that \emph{no} feedback (namely, the payoff $r_{t,a}$) is observed for \emph{unchosen} arms $a \ne a_t$.  The consequence of this fact is discussed in more details in the next subsection.}\label{def:bandit-feedback}
\end{compactenum}

In the process above, the \emph{total $T$-trial payoff} of $\Aalg$ is defined as
$
\sum_{t=1}^T r_{t,a_t}.
$
Similarly, we define the \emph{optimal expected $T$-trial payoff} as
$
\E\left[\sum_{t=1}^Tr_{t,a_t^*}\right],
$
where $a_t^*$ is the arm with maximum expected payoff at trial $t$.
Our goal is to design $\Aalg$ so that the expected total payoff
above is maximized.  Equivalently, we may find an algorithm
so that its \emph{regret} with respect to the optimal arm-selection
strategy is minimized.  Here, the $T$-trial regret $R_\Aalg(T)$ of
algorithm $\Aalg$ is defined formally by
\begin{eqnarray}
R_\Aalg(T) &\defeq& \E\left[\sum_{t=1}^Tr_{t,a_t^*}\right] - \E\left[\sum_{t=1}^Tr_{t,a_t}\right]. \label{eqn:regret-def}
\end{eqnarray}

An important special case of the general contextual bandit problem is the well-known \emph{$K$-armed bandit} in which (i) the arm set $\Aset_t$ remains unchanged and contains $K$ arms for all $t$, and (ii) the user $u_t$ (or equivalently, the context $(\vecb{x}_{t,1},\cdots,\vecb{x}_{t,K})$) is the same for all $t$.  Since both the arm set and contexts are constant at every trial, they make no difference to a bandit algorithm, and so we will also refer to this type of bandit as a \emph{context-free} bandit.

In the context of article recommendation, we may view articles in the pool as arms.  When a presented article is clicked, a payoff of $1$ is incurred; otherwise, the payoff is $0$.  With this definition of payoff, the expected payoff of an article is precisely its \emph{click-through rate (CTR)}, and choosing an article with maximum CTR is equivalent to maximizing the expected number of clicks from users, which in turn is the same as maximizing the total expected payoff in our bandit formulation.

Furthermore, in web services we often have access to user information which can be used to infer a user's interest and to choose news articles that are probably most interesting to her.
For example, it is much more likely for a male teenager to be interested in
an article about iPod products rather than retirement plans.
Therefore, we may ``summarize'' users and articles by a set of informative features that describe them compactly.  By doing so, a bandit algorithm can \emph{generalize} CTR
information from one article/user to another,
and learn to choose good articles more quickly, especially for new users and articles.

\subsection{Existing Bandit Algorithms} \label{sec:existing-algorithms}

The fundamental challenge in bandit problems is the need for balancing
exploration and exploitation.  To minimize the regret in
\eqnref{eqn:regret-def}, an algorithm $\Aalg$ \emph{exploits} its
past experience to select the arm that appears best.  On the other
hand, this seemingly optimal arm may in fact be suboptimal, due to
imprecision in $\Aalg$'s knowledge.  In order to avoid this undesired
situation, $\Aalg$ has to \emph{explore} by actually choosing
seemingly suboptimal arms so as to gather more information about them (\cf,
step~\ref{def:bandit-feedback} in the bandit process defined in the previous subsection).
Exploration can increase \emph{short-term} regret since some
suboptimal arms may be chosen.  However, obtaining information about
the arms' average payoffs (\ie, exploration) can refine $\Aalg$'s
estimate of the arms' payoffs and in turn reduce \emph{long-term} regret.
Clearly, neither a purely exploring nor a purely exploiting algorithm works
best in general, and a good tradeoff is needed.

The context-free $K$-armed bandit problem has been studied by statisticians
for a long time~\cite{Berry85Bandit,Robbins52Some,Thompson33Likelihood}.
One of the simplest and most straightforward algorithms is \algfont{$\epsilon$-greedy}.
In each trial $t$, this algorithm first estimates the average payoff $\hat{\mu}_{t,a}$ of each arm $a$.  Then, with probability $1-\epsilon$, it chooses the \emph{greedy} arm (\ie, the arm with highest payoff estimate); with probability $\epsilon$, it chooses a random arm.  In the limit, each arm will be tried infinitely often, and so the payoff estimate $\hat{\mu}_{t,a}$ converges to the true value $\mu_a$ with probability $1$.  Furthermore, by decaying $\epsilon$ appropriately (\eg, \cite{Robbins52Some}), the per-step regret, $R_\Aalg(T)/T$, converges to $0$ with probability $1$.

In contrast to the \emph{unguided} exploration strategy adopted by \algfont{$\epsilon$-greedy}, another class of algorithms generally known as upper confidence bound algorithms~\cite{Agrawal95Sample,Auer02Finite,Lai85Asymptotically} use a smarter way to balance exploration and exploitation.  Specifically, in trial $t$, these algorithms estimate both the mean payoff $\hat{\mu}_{t,a}$ of each arm $a$ as well as a corresponding confidence interval $c_{t,a}$, so that $\abs{\hat{\mu}_{t,a}-\mu_a}<c_{t,a}$ holds with high probability.  They then select the arm that achieves a highest upper confidence bound (UCB for short): $a_t = \arg\max_a\left(\hat{\mu}_{t,a}+c_{t,a}\right)$.  With appropriately defined confidence intervals, it can be shown that such algorithms have a small total $T$-trial regret that is only logarithmic in the total number of trials $T$, which turns out to be optimal~\cite{Lai85Asymptotically}.

While context-free $K$-armed bandits are extensively studied and well understood, the more general contextual bandit problem has remained challenging.  The \algfont{EXP4} algorithm~\cite{Auer02Nonstochastic} uses the exponential weighting technique to achieve an $\tilde{O}(\sqrt{T})$ regret,\footnote{Note $\tilde{O}(\cdot)$ is the same as 
$O(\cdot)$ but suppresses logarithmic factors.} but the computational complexity may be exponential in the number of features.  Another general contextual bandit algorithm is the \algfont{epoch-greedy} algorithm~\cite{Langford08Epoch} that is similar to \algfont{$\epsilon$-greedy} with shrinking $\epsilon$.  This algorithm is computationally efficient given an oracle optimizer but has the weaker regret guarantee of $\tilde{O}(T^{2/3})$.

Algorithms with stronger regret guarantees may be designed under various modeling assumptions about the bandit.  Assuming the expected payoff of an arm is linear in its features, Auer~\cite{Auer02Using} describes the \algfont{LinRel} algorithm that is essentially a UCB-type approach and shows that one of its variants has a regret of $\tilde{O}(\sqrt{T})$, a significant improvement over earlier algorithms~\cite{Abe03Reinforcement}.

Finally, we note that there exist another class of bandit algorithms
based on Bayes rule, such as Gittins index methods~\cite{Gittens79}.  With
appropriately defined prior distributions, Bayesian approaches may
have good performance.  These methods require extensive offline
engineering to obtain good prior models, and are often computationally
prohibitive without coupling with approximation techniques~\cite{Agarwal09Explore}.

\section{Algorithm} \label{sec:linucb}

Given asymptotic optimality and the strong regret bound of UCB methods for context-free bandit algorithms, it is tempting to devise similar algorithms for contextual bandit problems.  Given some parametric form of payoff function, a number of methods exist to estimate from data the confidence interval of the parameters with which we can compute a UCB of the estimated arm payoff.  Such an approach, however, is expensive in general.

In this work, we show that a confidence interval can be computed \emph{efficiently in closed form} when the payoff model is linear, and call this algorithm \algfont{LinUCB}.  For convenience of exposition, we first describe the simpler form for \emph{disjoint} linear models, and then consider the general case of \emph{hybrid} models in \secref{sec:linucb-hybrid}.  We note \algfont{LinUCB}
is a generic contextual bandit algorithms which applies to applications other than personalized news article recommendation.

\subsection{LinUCB with Disjoint Linear Models} \label{sec:linucb-disjoint}

Using the notation of \secref{sec:bandit}, we assume the expected payoff of an arm $a$ is linear in its $d$-dimensional feature $\vecb{x}_{t,a}$ with some unknown coefficient vector $\pmb{\theta}_a^*$; namely, for all $t$,
\begin{eqnarray}
\E[r_{t,a}|\vecb{x}_{t,a}] &=& \vecb{x}_{t,a}^\mt\pmb{\theta}_a^*. \label{eqn:model-disjoint}
\end{eqnarray}
This model is called \emph{disjoint} since the parameters are not
shared among different arms.
Let $\vecb{D}_a$ be a design matrix of dimension $m \times d$ at trial $t$, whose rows correspond to
$m$ training inputs (\eg, $m$ contexts that are observed previously for article $a$), and $\vecb{b}_a\in\Rset^m$ be the corresponding response vector (\eg, the corresponding $m$ click/no-click user feedback).  Applying ridge regression to the training data $(\vecb{D}_a,\vecb{c}_a)$ gives an estimate of the coefficients:
\begin{eqnarray}
\hat{\pmb{\theta}}_a = (\vecb{D}_a^\mt\vecb{D}_a+\vecb{I}_d)^\mi\vecb{D}_a^\mt\vecb{c}_a, \label{eqn:ridge-regression}
\end{eqnarray}
where $\vecb{I}_d$ is the $d\times d$ identity matrix.  When components in $\vecb{c}_a$ are independent conditioned on corresponding rows in $\vecb{D}_a$, it can be shown~\cite{Walsh09Exploring} %
that, with probability at least $1-\delta$,
\begin{eqnarray}
\abs{\vecb{x}_{t,a}^\mt\hat{\pmb{\theta}}_a-\E[r_{t,a}|\vecb{x}_{t,a}]} \le \alpha \sqrt{\vecb{x}_{t,a}^\mt(\vecb{D}_a^\mt\vecb{D}_a+\vecb{I}_d)^\mi\vecb{x}_{t,a}} \label{eqn:linucb-error-bound}
\end{eqnarray}
for any $\delta>0$ and $\vecb{x}_{t,a}\in\Rset^d$, where
$\alpha=1+\sqrt{\ln(2/\delta)/2}$ is a constant. In other words,
the inequality above gives a reasonably tight UCB for the expected
payoff of arm $a$, from which a UCB-type arm-selection strategy can be derived: at each trial $t$, choose
\begin{eqnarray}
a_t \defeq \arg\max_{a\in\Aset_t} \left(\vecb{x}_{t,a}^\mt\hat{\pmb{\theta}}_a+\alpha
\sqrt{\vecb{x}_{t,a}^\mt \vecb{A}_a^\mi\vecb{x}_{t,a}}\right), \label{eqn:criterion}
\end{eqnarray}
where $\vecb{A}_a \defeq \vecb{D}_a^\mt\vecb{D}_a+\vecb{I}_d$.

The confidence interval in \eqnref{eqn:linucb-error-bound} may be
motivated and derived from other principles. For instance,
ridge regression can also be interpreted as a Bayesian point
estimate, where the posterior distribution of the coefficient vector,
denoted as $p(\pmb{\theta}_a)$, is Gaussian with mean
$\hat{\pmb{\theta}}_a$ and covariance $\vecb{A}_a^\mi$.
Given the current model, the predictive variance of the expected
payoff $\vecb{x}_{t,a}^\mt\pmb{\theta}_a^*$ is evaluated as
$\vecb{x}_{t,a}^\mt\vecb{A}_a^\mi\vecb{x}_{t,a}$, and then
$\sqrt{\vecb{x}_{t,a}^\mt\vecb{A}_a^\mi\vecb{x}_{t,a}}$ becomes the standard
deviation.  Furthermore, in information
theory~\cite{MacKay03Information}, the differential entropy of
$p(\pmb{\theta}_a)$ is defined as $-\frac{1}{2}\ln
((2\pi)^d\det{\vecb{A}_a})$. The entropy of
$p(\pmb{\theta}_a)$ when updated by the inclusion of the new point $\vecb{x}_{t,a}$
then becomes $-\frac{1}{2}\ln
((2\pi)^d\det{(\vecb{A}_a+\vecb{x}_{t,a}\vecb{x}_{t,a}^\mt)})$. The entropy
reduction in the model posterior is $\frac{1}{2}\ln
(1+\vecb{x}_{t,a}^\mt\vecb{A}_a^\mi\vecb{x}_{t,a})$. This quantity is often
used to evaluate model improvement contributed from $\vecb{x}_{t,a}$.
Therefore, the criterion for arm selection in \eqnref{eqn:criterion} can
also be regarded as an additive trade-off between the payoff estimate
and model uncertainty reduction.

\algref{alg:linucb-disjoint} gives a detailed description of the entire \algfont{LinUCB} algorithm, whose only input parameter is $\alpha$.  Note the value of $\alpha$ given in \eqnref{eqn:linucb-error-bound} may be conservatively large in some applications, and so optimizing this parameter may result in higher total payoffs in practice.  Like all UCB methods, \algfont{LinUCB} always chooses the arm with highest UCB (as in \eqnref{eqn:criterion}).

This algorithm has a few nice properties.  First, its computational
complexity is linear in the number of arms and at most cubic in the
number of features.  To decrease computation further, we may update
$\vecb{A}_{a_t}$ in every step (which takes $O(d^2)$ time), but
compute and cache $\vecb{Q}_a\defeq\vecb{A}_a^\mi$ (for all $a$)
periodically instead of in real-time.  Second, the algorithm works
well for a dynamic arm set, and remains efficient as long as the size
of $\Aset_t$ is not too large.  This case is true in many
applications.  In news article recommendation, for instance, editors
add/remove articles to/from a pool and the pool size remains
essentially constant.  Third, although it is not the focus of the
present paper, we can adapt the analysis from \cite{Auer02Using} to
show the following: if the arm set $\Aset_t$ is fixed and contains $K$
arms, then the confidence interval (\ie, the right-hand side of
\eqnref{eqn:linucb-error-bound}) decreases fast enough with more and
more data, and then prove the strong regret bound of
$\tilde{O}(\sqrt{KdT})$, matching the state-of-the-art result~\cite{Auer02Using}
for bandits satisfying \eqnref{eqn:model-disjoint}.
These theoretical results indicate fundamental
soundness and efficiency of the algorithm.

Finally, we note that, under the assumption that input features $\vecb{x}_{t,a}$ were drawn i.i.d. from a normal distribution (in addition to the modeling assumption in \eqnref{eqn:model-disjoint}), Pavlidis \textit{et~al.}~\cite{Pavlidis08Simulation}
came up with a similar algorithm that uses a least-squares solution $\tilde{\pmb{\theta}}_a$ instead of our ridge-regression solution ($\hat{\pmb{\theta}}_a$ in \eqnref{eqn:ridge-regression}) to compute the UCB.  However, our approach (and theoretical analysis) is more general and remains valid even when input features are nonstationary.  More importantly, we will discuss in the next section how to extend the basic \algref{alg:linucb-disjoint} to a much more interesting case not covered by Pavlidis \textit{et~al}.

\begin{algorithm}[t]
\begin{algorithmic}[1]
\item Inputs: $\alpha\in\Rset_+$
\FOR{$t=1,2,3,\ldots,T$}
\STATE Observe features of all arms $a\in\Aset_t$: $\vecb{x}_{t,a}\in\Rset^d$
\FORALL{$a\in\Aset_t$}
\IF{$a$ is new}
\STATE $\vecb{A}_a \leftarrow \vecb{I}_d$ ($d$-dimensional identity matrix)
\STATE $\vecb{b}_a \leftarrow \mathbf{0}_{d\times1}$ ($d$-dimensional zero vector)
\ENDIF
\STATE $\hat{\pmb{\theta}}_a \leftarrow \vecb{A}_a^\mi\vecb{b}_a$
\STATE $p_{t,a} \leftarrow \hat{\pmb{\theta}}_a^\mt \vecb{x}_{t,a} +\alpha\sqrt{\vecb{x}_{t,a}^\mt\vecb{A}_a^\mi\vecb{x}_{t,a}}$ %
\ENDFOR
\STATE Choose arm $a_t = \arg\max_{a\in\Aset_t}p_{t,a}$ with ties broken arbitrarily,
and observe a real-valued payoff $r_t$
\STATE $\vecb{A}_{a_t} \leftarrow \vecb{A}_{a_t} + \vecb{x}_{t,a_t}\vecb{x}_{t,a_t}^\mt$
\STATE $\vecb{b}_{a_t} \leftarrow \vecb{b}_{a_t} + r_t\vecb{x}_{t,a_t}$
\ENDFOR
\end{algorithmic}
\caption{LinUCB with disjoint linear models.} \label{alg:linucb-disjoint}
\end{algorithm}

\subsection{LinUCB with Hybrid Linear Models} \label{sec:linucb-hybrid}

\algref{alg:linucb-disjoint} (or the similar algorithm in \cite{Pavlidis08Simulation}) computes the inverse of the matrix, $\vecb{D}_a^\mt\vecb{D}_a+\vecb{I}_d$ (or $\vecb{D}_a^\mt\vecb{D}_a$), where $\vecb{D}_a$ is again the design matrix with rows corresponding to features in the training data.  These matrices of all arms have fixed dimension $d\times d$, and can be updated efficiently and incrementally.  Moreover, their inverses can be computed easily as the parameters in \algref{alg:linucb-disjoint} are \emph{disjoint}: the solution $\hat{\pmb{\theta}}_a$ in \eqnref{eqn:ridge-regression} is not affected by training data of other arms, and so can be computed separately.  We now consider the more interesting case with \emph{hybrid} models.

In many applications including ours, it is helpful to use features
that are shared by all arms, in addition to the arm-specific 
ones.  For example, in news article recommendation, a user may
prefer only articles about politics for which this provides a
mechanism.  Hence, it is helpful to have features that
have both shared and non-shared components.
Formally, we adopt the following \emph{hybrid model} by
adding another linear term to the right-hand side of \eqnref{eqn:model-disjoint}:
\begin{eqnarray}
\E[r_{t,a}|\vecb{x}_{t,a}] &=& \vecb{z}_{t,a}^\mt\pmb{\beta}^* + \vecb{x}_{t,a}^\mt\pmb{\theta}_a^*, \label{eqn:model-hybrid}
\end{eqnarray}
where $\vecb{z}_{t,a}\in\Rset^k$ is the feature of the current user/article combination, and $\pmb{\beta}^*$ is an unknown coefficient vector common to all arms.
This model is hybrid in the sense that some of the coefficients $\pmb{\beta}^*$ are shared by all arms, while others $\pmb{\theta}_a^*$ are not.

For hybrid models, we can no longer use
\algref{alg:linucb-disjoint} as the confidence intervals of various
arms are not independent due to the shared features.  Fortunately,
there is an efficient way to compute an UCB along the same line of
reasoning as in the previous section.  The derivation relies heavily
on block matrix inversion techniques.  Due to space limitation, we
only give the pseudocode in
\algref{alg:linucb-hybrid} (where lines~\ref{alg:linucb-hybrid:beta} and \ref{alg:linucb-hybrid:theta} compute the ridge-regression solution of the coefficients, and line~\ref{alg:linucb-hybrid:ci} computes the confidence interval), and leave detailed derivations to a full
paper.  Here, we only point out the important fact that the algorithm
is computationally efficient since the building blocks in the
algorithm ($\vecb{A}_0$, $\vecb{b}_0$, $\vecb{A}_a$, $\vecb{B}_a$, and
$\vecb{b}_a$) all have fixed dimensions and can be updated
incrementally.  Furthermore, quantities associated with arms not existing
in $\Aset_t$ no longer get involved in the computation.
Finally, we can also compute and cache the
inverses ($\vecb{A}_0^\mi$ and $\vecb{A}_a^\mi$) periodically instead
of at the end of each trial to reduce the per-trial computational complexity
to $O(d^2+k^2)$.

\begin{algorithm}[t]
\begin{algorithmic}[1]
\item Inputs: $\alpha\in\Rset_+$
\STATE $\vecb{A}_0 \leftarrow \vecb{I}_k$ ($k$-dimensional identity matrix)
\STATE $\vecb{b}_0 \leftarrow \vecb{0}_k$ ($k$-dimensional zero vector)
\FOR{$t=1,2,3,\ldots,T$}
\STATE Observe features of all arms $a\in\Aset_t$: $(\vecb{z}_{t,a},\vecb{x}_{t,a})\in\Rset^{k+d}$
\STATE $\hat{\pmb{\beta}} \leftarrow \vecb{A}_0^\mi\vecb{b}_0$ \label{alg:linucb-hybrid:beta}
\FORALL{$a\in\Aset_t$}
\IF{$a$ is new}
\STATE $\vecb{A}_a \leftarrow \vecb{I}_d$ ($d$-dimensional identity matrix)
\STATE $\vecb{B}_a \leftarrow \vecb{0}_{d\times k}$ ($d$-by-$k$ zero matrix)
\STATE $\vecb{b}_a \leftarrow \mathbf{0}_{d\times1}$ ($d$-dimensional zero vector)
\ENDIF
\STATE $\hat{\pmb{\theta}}_a \leftarrow \vecb{A}_a^\mi\left(\vecb{b}_a-\vecb{B}_a\hat{\pmb{\beta}}\right)$ \label{alg:linucb-hybrid:theta}
\STATE $s_{t,a} \leftarrow \vecb{z}_{t,a}^\mt\vecb{A}_0^\mi\vecb{z}_{t,a} - 2\vecb{z}_{t,a}^\mt\vecb{A}_0^\mi\vecb{B}_a^\mt\vecb{A}_a^\mi\vecb{x}_{t,a} + \vecb{x}_{t,a}^\mt\vecb{A}_a^\mi\vecb{x}_{t,a} + \vecb{x}_{t,a}^\mt\vecb{A}_a^\mi\vecb{B}_a\vecb{A}_0^\mi\vecb{B}_a^\mt\vecb{A}_a^\mi\vecb{x}_{t,a}$ \label{alg:linucb-hybrid:ci}
\STATE $p_{t,a} \leftarrow \vecb{z}_{t,a}^\mt\hat{\pmb{\beta}} + \vecb{x}_{t,a}^\mt\hat{\pmb{\theta}}_a + \alpha \sqrt{s_{t,a}}$
\ENDFOR
\STATE Choose arm $a_t = \arg\max_{a\in\Aset_t}p_{t,a}$ with ties broken arbitrarily,
and observe a real-valued payoff $r_t$
\STATE $\vecb{A}_0 \leftarrow \vecb{A}_0 + \vecb{B}_{a_t}^\mt\vecb{A}_{a_t}^\mi\vecb{B}_{a_t}$
\STATE $\vecb{b}_0 \leftarrow \vecb{b}_0 + \vecb{B}_{a_t}^\mt\vecb{A}_{a_t}^\mi\vecb{b}_{a_t}$
\STATE $\vecb{A}_{a_t} \leftarrow \vecb{A}_{a_t} + \vecb{x}_{t,a_t}\vecb{x}_{t,a_t}^\mt$
\STATE $\vecb{B}_{a_t} \leftarrow \vecb{B}_{a_t} + \vecb{x}_{t,a_t} \vecb{z}_{t,a_t}^\mt$
\STATE $\vecb{b}_{a_t} \leftarrow \vecb{b}_{a_t} + r_t\vecb{x}_{t,a_t}$
\STATE $\vecb{A}_0 \leftarrow \vecb{A}_0 + \vecb{z}_{t,a_t}\vecb{z}_{t,a_t}^\mt - \vecb{B}_{a_t}^\mt\vecb{A}_{a_t}^\mi\vecb{B}_{a_t}$
\STATE $\vecb{b}_0 \leftarrow \vecb{b}_0 + r_t\vecb{z}_{t,a_t} - \vecb{B}_{a_t}^\mt\vecb{A}_{a_t}^\mi\vecb{b}_{a_t}$
\ENDFOR
\end{algorithmic}
\caption{LinUCB with hybrid linear models.} \label{alg:linucb-hybrid}
\end{algorithm}

\section{Evaluation Methodology} \label{sec:evaluation}

Compared to machine learning in the more standard supervised setting,
evaluation of methods in a contextual bandit setting is frustratingly
difficult.
Our goal here is to measure the performance of a {\em bandit algorithm} $\pi$,
that is, a rule for selecting an arm at each time step based on the
preceding interactions (such as the algorithms described above).
Because of the interactive nature of the problem, it would seem that
the only way to do this is to actually run the algorithm on ``live''
data.
However, in practice, this approach is likely to be infeasible due to the
serious logistical challenges that it presents.
Rather, we may only
have {\em offline} data available that was collected at a previous
time using an entirely {\em different} logging policy.
Because payoffs are only observed for the arms chosen by the logging
policy, which are likely to often differ from those chosen by the
algorithm $\pi$ being evaluated, it is not at all clear how to evaluate
$\pi$ based only on such logged data.
This evaluation problem may be viewed as a special case of the so-called
``off-policy evaluation problem'' in reinforcement learning (see, \textit{c.f.},
\cite{Precup00Eligibility}).

One solution is to build
a simulator to model the bandit process from the logged data, and then
evaluate $\pi$ with the simulator.  However, the modeling step will
introduce \emph{bias} in the simulator and so make it hard to justify the
reliability of this simulator-based evaluation approach.
In contrast, we propose an approach that is simple to implement,
grounded on logged data, and \emph{unbiased}.

In this section, we describe a provably reliable technique for carrying out
such an evaluation, assuming that the individual events are
i.i.d., and that the logging
policy that was used to gather the logged data chose each arm at each
time step uniformly at random.
Although we omit the details, this latter assumption can be
weakened considerably so that any randomized logging policy is allowed
and our solution can be modified accordingly using rejection
sampling, but at the cost of decreased efficiency in using data.

More precisely, we suppose that there is some unknown distribution $D$
from which tuples are drawn i.i.d.~of the form
$(\vecb{x}_{1}, ..., \vecb{x}_{K}, r_{1}, \ldots, r_{K})$, each
consisting of observed feature vectors and \emph{hidden} payoffs for all arms.
We also posit access to a large sequence of logged events resulting
from the interaction of the logging policy with the world.
Each such event consists of the context vectors
$\vecb{x}_1,...,\vecb{x}_K$, a selected arm $a$ and the resulting
observed payoff $r_a$.
Crucially, only the payoff $r_a$ is observed for the single arm $a$
that was chosen uniformly at random.
For simplicity of presentation, we take this sequence of logged events
to be an infinitely long stream; however, we also give explicit bounds
on the actual finite number of events required by our evaluation method.

Our goal is to use this data to evaluate a bandit algorithm
$\pi$.  Formally, $\pi$ is a (possibly randomized) mapping
for selecting the arm $a_t$ at time $t$ based on the history $h_{t-1}$
of $t-1$ preceding events, together with the current context vectors
$\vecb{x}_{t1},...,\vecb{x}_{tK}$.

Our proposed policy evaluator is shown in \algref{alg:pe}.
The method takes as input a policy $\pi$ and a desired number of
``good'' events $T$ on which to base the evaluation.
We then step through the stream of logged events one by one.
If, given the current history $h_{t-1}$, it happens that the policy
$\pi$ chooses the same arm $a$ as the one that was selected by the
logging policy, then the event is retained, that is, added to the
history, and the total payoff $R_t$ updated.
Otherwise, if the policy $\pi$ selects a different arm from the one
that was taken by the logging policy, then the event is entirely
ignored, and the algorithm proceeds to the next event without any
other change in its state.

Note that, because the logging policy chooses each arm uniformly at
random, each event is retained by this algorithm with probability
exactly $1/K$, independent of everything else.
This means that the events which are retained have the same
distribution as if they were selected by $D$.
As a result, we can prove that two processes are
equivalent: the first is evaluating the policy against $T$ real-world
events from $D$, and the second is evaluating the policy using the policy
evaluator on a stream of logged events.

\newtheorem{theorem}{Theorem}
\begin{theorem} \label{thm:simulation}
For all distributions $D$ of contexts, %
all policies $\pi$, all $T$, and all sequences of events $h_T$,
\[ \Pr_{\mbox{Policy\_Evaluator}(\pi,S)}(h_T) = \Pr_{\pi,D} (h_T) \]
where $S$ is a stream of events drawn i.i.d. from a uniform random logging
policy and $D$.
Furthermore, the expected number of events obtained from the stream to
gather a history $h_T$ of length $T$ is $KT$.
\end{theorem}

This theorem says that {\em every} history $h_T$ has the identical probability
in the real world as in the policy evaluator.  Many statistics of these
histories, such as the average payoff $R_T/T$ returned by \algref{alg:pe},
are therefore unbiased estimates of the value of the algorithm $\pi$.
Further, the theorem states that $KT$ logged events are required, in
expectation, to retain a sample of size $T$.

\begin{proof}
The proof is by induction on $t=1,\ldots,T$ starting with a base case
of the empty
history which has probability $1$ when $t=0$ under both methods of
evaluation.  In the inductive case, assume that we have for all $t-1$:
\[ \Pr_{\mbox{Policy\_Evaluator}(\pi,S)}(h_{t-1}) = \Pr_{\pi,D}
(h_{t-1})\] and want to prove the same statement for any history
$h_t$.  Since the data is i.i.d. and any randomization in the policy is
independent of randomization in the world, we need only prove that
conditioned on the history $h_{t-1}$ the distribution over the $t$-th event is
the same for each process.  In other words, we must show:
\begin{align*}
& \Pr_{\mbox{Policy\_Evaluator}(\pi,S)}((\vecb{x}_{t,1},...,\vecb{x}_{t,K},a,r_{t,a}) \mid h_{t-1})\\
= & \Pr_{D}(\vecb{x}_{t,1},...,\vecb{x}_{t,K},r_{t,a})\Pr_{\pi(h_{t-1})}(a \mid\vecb{x}_{t,1},...,\vecb{x}_{t,K}).
\end{align*}
Since the arm $a$ is chosen uniformly at random in the logging
policy, the probability that the policy evaluator exits the inner loop
is identical for any policy, any history, any features, and any arm,
implying this happens for the last event with the
probability of the last event, $\Pr_{D}(\vecb{x}_{t,1},...,\vecb{x}_{t,K},r_{t,a})$.
Similarly, since the policy $\pi$'s distribution over arms is
independent conditioned on the history $h_{t-1}$ and features
$(\vecb{x}_{t,1},...,\vecb{x}_{t,K})$, the probability of arm $a$ is just
$\Pr_{\pi(h_{t-1})}(a|\vecb{x}_{t,1},...,\vecb{x}_{t,K})$.

Finally,
since each event from the stream is retained with probability exactly
$1/K$, the expected number required to retain $T$ events is exactly $KT$.
\end{proof}

\begin{algorithm}[t]
\begin{algorithmic}[1]
\item Inputs: $T>0$; policy $\pi$; stream of events
\STATE $h_0\leftarrow\emptyset$ \COMMENT{An initially empty history}
\STATE $R_0\leftarrow0$ \COMMENT{An initially zero total payoff}
\FOR{$t=1,2,3,\ldots,T$}
\REPEAT
\STATE Get next event $(\vecb{x}_1,...,\vecb{x}_K,a,r_a)$
\UNTIL{$\pi(h_{t-1},(\vecb{x}_{1},...,\vecb{x}_{K})) = a$}
\STATE $h_t \leftarrow \mbox{\sc concatenate}(h_{t-1}, (\vecb{x}_{1},...,\vecb{x}_{K},a,r_{a}))$
\STATE $R_t \leftarrow R_{t-1} + r_{a}$
\ENDFOR
\STATE Output: $R_{T}/T$
\end{algorithmic}
\caption{Policy\_Evaluator.} \label{alg:pe}
\end{algorithm}

\section{Experiments}\label{sec:experiments}

In this section, we verify the capacity of the proposed \algfont{LinUCB}
algorithm on a real-world application using the offline evaluation
method of \secref{sec:evaluation}. We start with an introduction
of the problem setting in Yahoo!\ Today-Module, and then describe
the user/item attributes we used in experiments. Finally, we
define performance metrics and report experimental results with
comparison to a few standard (contextual) bandit algorithms.

\begin{figure}
\centering
\includegraphics[width=2.95in]{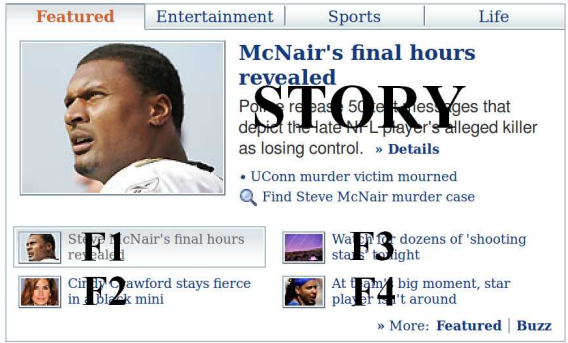}
\figsqueeze
\caption{A snapshot of the ``Featured'' tab in the Today Module on
Yahoo!\ Front Page. By default, the article at F1 position is
highlighted
at the story position.} \label{fig:fptoday}%
\end{figure}

\subsection{Yahoo!\ Today Module}

The Today Module is the most prominent panel on the Yahoo!\ Front Page,
which is also one of the most visited pages on the Internet; see a
snapshot in Figure \ref{fig:fptoday}. The default ``Featured'' tab in
the Today Module highlights one of four high-quality articles, mainly
news, while the four articles are selected from an hourly-refreshed
article pool curated by human editors. As illustrated in Figure
\ref{fig:fptoday}, there are four articles at footer positions,
indexed by F1--F4.  Each article is
represented by a small picture and a title. One of the four articles
is highlighted at the story position, which is featured by a large
picture, a title and a short summary along with related
links. {By default, the article at F1 is highlighted at the story
  position.} A user can click on the highlighted article at the story
position to read more details if she is interested in the article. The
event is recorded as a story click. \comment{If a user is interested
  in an article at F2$\sim$F4 positions, she can highlight the article
  at the story position by clicking on the respective footer
  position. }To draw visitors' attention, we would like to rank
available articles according to individual interests, and highlight
the most attractive article for each visitor at the story position.

\begin{figure*}[t]
\begin{center}
\subfigure[Deployment bucket.]{\includegraphics[angle=0,width=0.85\columnwidth]{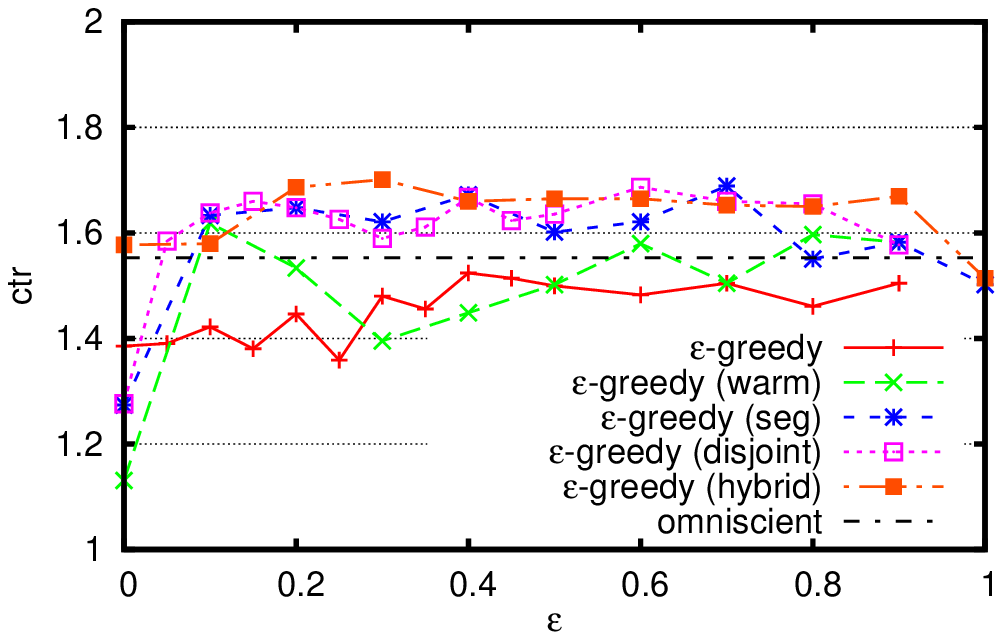} \label{fig:vd-expctr-allegd}} 
\subfigure[Deployment bucket.]{\includegraphics[angle=0,width=0.85\columnwidth]{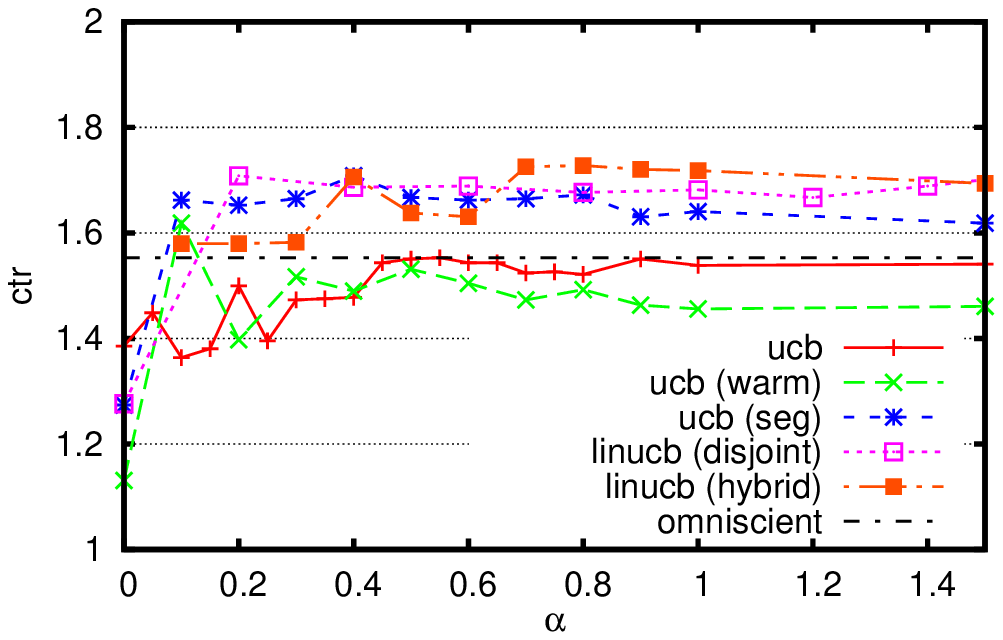} \label{fig:vd-expctr-allucb}}
\subfigure[Learning bucket.]{\includegraphics[angle=0,width=0.85\columnwidth]{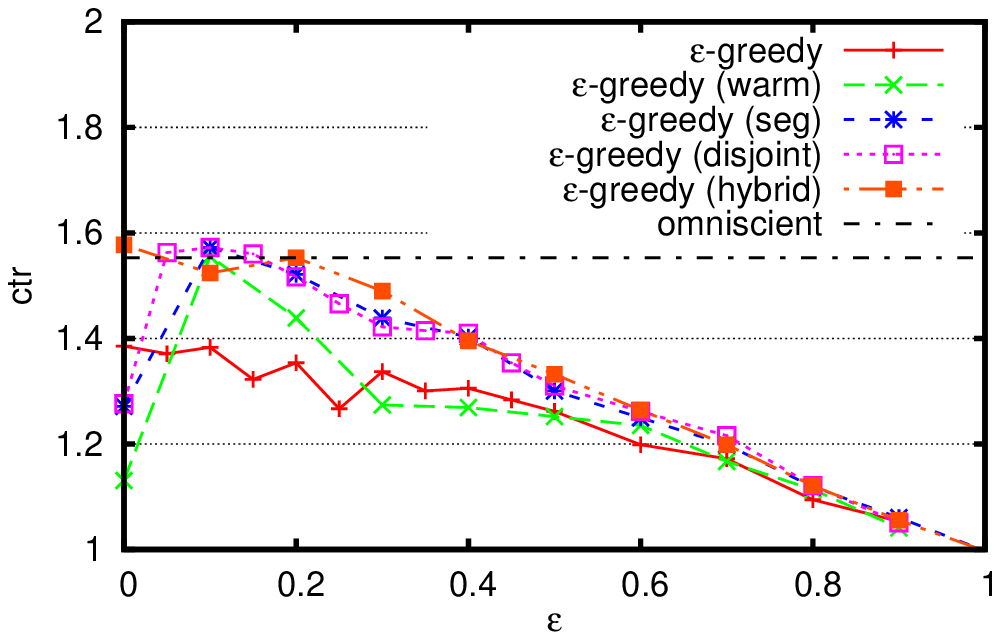} \label{fig:vd-ovrctr-allegd}}
\subfigure[Learning bucket.]{\includegraphics[angle=0,width=0.85\columnwidth]{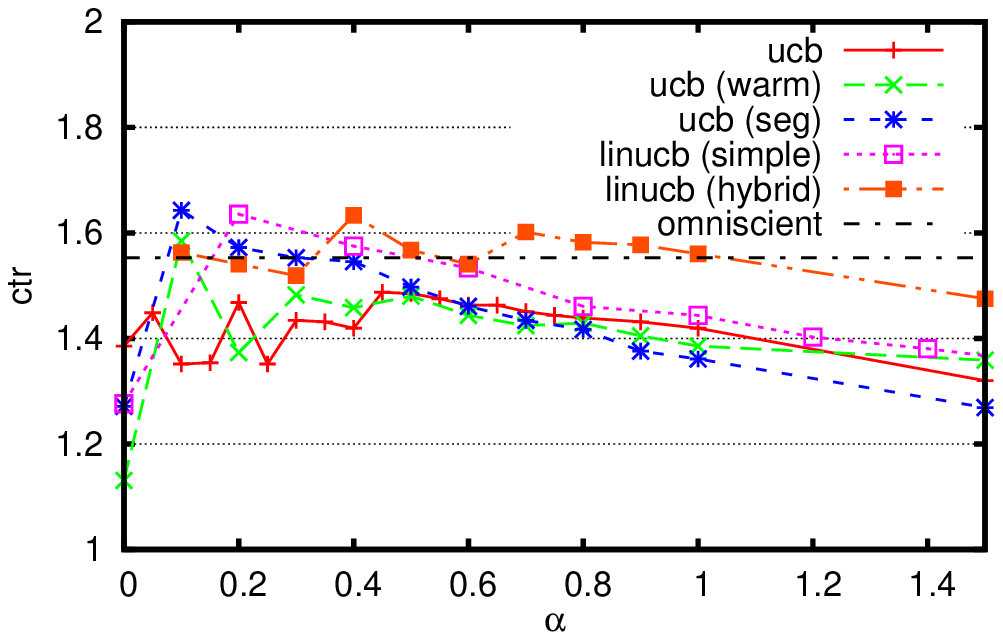} \label{fig:vd-ovrctr-allucb}}
\end{center}
\figsqueeze
\caption{Parameter tuning: CTRs of various algorithms on the one-day tuning dataset.} \label{fig:vd-ctr-all}
\end{figure*}

\subsection{Experiment Setup}

\begin{table*}[t]
\begin{center}
{\small
\input{td-ctr-all.tex}
}
\end{center}
\figsqueeze
\caption{Performance evaluation: CTRs of all algorithms on the one-week evaluation dataset in the deployment and learning buckets (denoted by ``deploy'' and ``learn'' in the table, respectively).  The numbers with a percentage is the CTR lift compared to \algfont{$\epsilon$-greedy}.} \label{tbl:td-ctr}
\end{table*}

This subsection gives a detailed description of our experimental
setup, including data collection, feature construction,
performance evaluation, and competing algorithms.

\subsubsection{Data Collection} \label{sec:data}

We collected events from a random bucket in May 2009. Users were
randomly selected to the bucket with a certain probability
per visiting view.\footnote{We call it view-based randomization. After
  refreshing her browser, the user may not fall into the random bucket
  again.} In this bucket, articles were randomly
selected from the article pool to serve users. To avoid exposure bias
at footer positions, we only focused on users' interactions with
F1 articles at the story position.  Each user interaction \emph{event}
consists of three components: (i) the random article chosen to serve
the user, (ii) user/article information, and (iii) whether the user
clicks on the article at the story position.  
\secref{sec:evaluation} shows these random events can be used to reliably evaluate a
bandit algorithm's expected payoff.

There were about $4.7$ million events in the random bucket on May
01. We used this day's events (called ``tuning data'') for model validation
to decide the
optimal parameter for each competing bandit algorithm.  Then we ran
these algorithms with tuned parameters on a one-week event set (called
``evaluation data'') in the random bucket from May 03--09, which contained
about $36$ million events.

\subsubsection{Feature Construction} \label{sec:features}

We now describe the user/article features constructed for our
experiments.  Two sets of features for the disjoint and hybrid models, respectively,
were used to test the two forms of \algfont{LinUCB} in
\secref{sec:linucb} and to verify our conjecture that hybrid
models can improve learning speed.

We start with raw user features that were selected by ``support''.  The support of a feature is the fraction of users having that feature.  To reduce noise in the data, we only selected features with high support.  Specifically, we used a feature when its support is at least $0.1$.  %
Then, each user was originally represented by a raw feature vector
of over $1000$ categorical components, which include: (i)
demographic information: gender ($2$ classes) and age discretized
into $10$ segments; (ii) geographic features: about $200$
metropolitan locations worldwide and U.S. states; and (iii)
behavioral categories: about $1000$ binary categories that
summarize the user's consumption history within Yahoo!\ properties.
Other than these features, no other information was used
to identify a user.

Similarly, each article was represented by a raw feature vector of
about $100$ categorical features constructed in the same way.
These features include: (i) URL categories: tens of classes
inferred from the URL of the article resource; and (ii) editor
categories: tens of topics tagged by human editors to summarize
the article content.

We followed a previous procedure~\cite{Chu09Personalized}
to encode categorical user/article features as binary vectors and
then normalize each feature vector to unit length.  We also
augmented each feature vector with a constant feature of value $1$.  Now each
article and user was represented by a feature vector of $83$ and
$1193$ entries, respectively.

\comment{For example, ``gender'' of two classes is translated into
two binary features, i.e., ``male'' is encoded as $[0,1]$,
``female'' is encoded as $[1,0]$ and ``unknown'' is
$[0,0]$.\footnote{The ``unknown'' category coded with zero entries
has little contribution to our linear models.} As the number of
non-zero entries in these binary feature vectors varies, we
further normalized each vector into unit length, i.e., non-zero
entries in the normalized vector are replaced by $1/\sqrt{k}$,
where $k$ is the number of non-zero entries. For user features, we
normalized behavioral categories and the remaining features (age,
gender and location) separately, due to the variable length of
behavioral categories per user. For article features, we
normalized URL and Editor categories together. We also augmented
each feature vector by adding a constant term $1$. Each content
item is represented by a feature vector of $83$ entries, while
each user is represented by a feature vector of $1193$ entries.}

To further reduce dimensionality and capture nonlinearity in
these raw features, we carried out conjoint
analysis based on random exploration data collected in September
2008.  Following a previous approach to dimensionality
reduction~\cite{Chu09Conjoint}, we
projected user features onto article categories and then clustered
users with similar preferences into groups.  More specifically:
\begin{compactitem}
\item{We first used logistic regression (LR) to fit a bilinear model for
  click probability given raw user/article features so that
  $\pmb{\phi}_u^\mt \vecb{W} \pmb{\phi}_a$ approximated the
  probability that the user $u$ clicks on article $a$, where
  $\pmb{\phi}_u$ and $\pmb{\phi}_a$ were the corresponding feature
  vectors, and $\vecb{W}$ was a weight matrix optimized by LR.}
\item{Raw user features were then projected onto an induced space by computing $\pmb{\psi}_u \defeq \pmb{\phi}_u^\mt \vecb{W}$. Here, the $i^\mathrm{th}$ component in %
$\pmb{\psi}_u$ for user $u$ may be interpreted as the degree to which the user likes the $i^\mathrm{th}$ category of articles.  K-means was applied to group users in the induced $\pmb{\psi}_u$ space into $5$ clusters.}
\item{The final user feature was a six-vector: five entries corresponded to membership of that user in these $5$ clusters (computed with a Gaussian kernel
and then normalized so that they sum up to unity), and the sixth was a constant feature $1$.}
\end{compactitem}
At trial $t$, each article $a$ has a separate six-dimensional
feature $\vecb{x}_{t,a}$ that is exactly the six-dimensional
feature constructed as above for user $u_t$.  Since these article
features do not overlap, they are for disjoint linear models
defined in \secref{sec:linucb}.

For each article $a$, we performed the same dimensionality reduction
to obtain a six-dimensional article feature
(including a constant $1$ feature).  Its outer product with a user
feature gave $6 \times 6 = 36$ features, denoted $\vecb{z}_{t,a}\in\Rset^{36}$,
that corresponded to the shared features in \eqnref{eqn:model-hybrid},
and thus $(\vecb{z}_{t,a},\vecb{x}_{t,a})$ could be used in the hybrid linear model.
Note the features $\vecb{z}_{t,a}$ contains user-article interaction information,
while $\vecb{x}_{t,a}$ contains user information only.

Here, we intentionally used five users (and articles) groups,
which has been shown to be representative in segmentation
analysis~\cite{Chu09Conjoint}.  Another reason for using a relatively
small feature space is that, in online services, storing and retrieving
large amounts of user/article information will be too expensive to be practical.

\subsection{Compared Algorithms} \label{sec:algorithms}

The algorithms empirically evaluated in our experiments can be
categorized into three groups:

\noindent \textbf{I.~Algorithms that make no use of features.}  These
correspond to the context-free $K$-armed bandit algorithms
that ignore all contexts (\ie, user/article information).
\begin{compactitem}
\item{\algfont{random}: A random policy always chooses one of the candidate articles from the pool with equal probability.  This algorithm requires no parameters and does not ``learn'' over time.}
\item{\algfont{$\epsilon$-greedy}: As described in \secref{sec:existing-algorithms}, it estimates each article's CTR; then it chooses a random article with probability $\epsilon$, and chooses the article of the highest CTR estimate with probability $1-\epsilon$.  The only parameter of this policy is $\epsilon$.}
\item{\algfont{ucb}: As described in \secref{sec:existing-algorithms}, this policy estimates each article's CTR as well as a confidence interval of the estimate, and always chooses the article with the highest UCB.  Specifically, following \algfont{UCB1}~\cite{Auer02Finite},
we computed an article $a$'s confidence interval by $c_{t,a}=\frac{\alpha}{\sqrt{n_{t,a}}}$, where $n_{t,a}$ is the number of times $a$ was chosen prior to trial $t$, and $\alpha>0$ is a parameter.}
\item{\algfont{omniscient}: Such a policy achieves the best empirical context-free CTR from \emph{hindsight}.  It first computes each article's empirical CTR from logged events, and then always chooses the article with highest empircal CTR when it is evaluated using the \emph{same} logged events.  This algorithm requires no parameters and does not ``learn'' over time.}
\end{compactitem}

\noindent \textbf{II.~Algorithms with ``warm start''}---an intermediate step
towards personalized services. %
The idea is to provide an offline-estimated user-specific
adjustment on articles' context-free CTRs over the whole traffic. The offset
serves as an initialization on CTR estimate for new content,
\aka ``warm start''. We re-trained the bilinear logistic regression
model studied in \cite{Chu09Personalized} on Sept 2008 random traffic
data, using features $\vecb{z}_{t,a}$ constructed above. The selection
criterion then becomes the sum of the context-free CTR estimate and
a bilinear term for a user-specific CTR adjustment.
In training, CTR was estimated using the context-free
\algfont{$\epsilon$-greedy} with $\epsilon=1$.

\begin{compactitem}
\item{\algfont{$\epsilon$-greedy (warm)}: This algorithm is the same as \algfont{$\epsilon$-greedy} except it adds the user-specific CTR correction to the article's context-free CTR estimate.}
\item{\algfont{ucb (warm)}: This algorithm is the same as the previous one but replaces \algfont{$\epsilon$-greedy} with \algfont{ucb}.}
\end{compactitem}

\noindent \textbf{III.~Algorithms that learn user-specific CTRs
online.}
\begin{compactitem}
\item{\algfont{$\epsilon$-greedy (seg)}: Each user is assigned to the closest user cluster among the five constructed in \secref{sec:features}, and so all users are partitioned into five groups (\aka\ user segments), in each of which a separate copy of \algfont{$\epsilon$-greedy} was run.}
\item{\algfont{ucb (seg)}: This algorithm is similar to \algfont{$\epsilon$-greedy (seg)} except it ran a copy of \algfont{ucb} in each of the five user segments.}
\item{\algfont{$\epsilon$-greedy (disjoint)}: This is \algfont{$\epsilon$-greedy} with disjoint models, and may be viewed as a close variant of \algfont{epoch-greedy}~\cite{Langford08Epoch}.}
\item{\algfont{linucb (disjoint)}: This is \algref{alg:linucb-disjoint} with disjoint models.}
\item{\algfont{$\epsilon$-greedy (hybrid)}: This is \algfont{$\epsilon$-greedy} with hybrid models, and may be viewed as a close variant of \algfont{epoch-greedy}.}
\item{\algfont{linucb (hybrid)}: This is \algref{alg:linucb-hybrid} with hybrid models.}
\end{compactitem} %

\subsection{Performance Metric}

An algorithm's CTR is defined as the ratio of the number of clicks
it receives and the number of steps it is run.
We used all algorithms' CTRs on the random logged events for
performance comparison.  To protect
business-sensitive information, we report an algorithm's \emph{relative
CTR}, which is the algorithm's CTR divided by the random policy's.
Therefore, we will not report a random policy's relative CTR as it
is always $1$ by definition.  For convenience, we will use the term
``CTR'' from now on instead of ``relative CTR''.

For each algorithm, we are interested in two CTRs motivated by our
application, which may be useful for other similar applications.  When deploying the methods to Yahoo!'s front page, one
reasonable way is to randomly split all traffic to this page into
two buckets~\cite{Agarwal09Online}.  The first, called ``learning bucket'', usually consists of
a small fraction of traffic on which various bandit algorithms are run to
learn/estimate article CTRs.  The other, called ``deployment bucket'',
is where Yahoo!\ Front Page greedily serves users using CTR estimates obained
from the learning bucket.  Note that ``learning'' and ``deployment'' are
interleaved in this problem, and so in every view falling into the deployment
bucket, the article with the highest \emph{current} (user-specific) CTR estimate
is chosen; this estimate may change later if the learning bucket gets more data.
CTRs in both buckets were estimated with \algref{alg:pe}.

Since the deployment bucket is often larger than the learning bucket,
CTR in the deployment bucket is more important.  However, a higher CTR in
the learning bucket suggests a faster learning rate (or equivalently, smaller
regret) for a bandit algorithm.  Therefore, we chose to
report algorithm CTRs in both buckets. \vspace{-1mm}

\subsection{Experimental Results} \label{sec:results} \vspace{-2mm}

\subsubsection{Results for Tuning Data} \label{sec:validation-results} \vspace{-1mm}

Each of the competing algorithms (except \algfont{random} and
\algfont{omniscient}) in \secref{sec:algorithms} requires a single
parameter: $\epsilon$ for $\epsilon$-greedy algorithms and $\alpha$
for UCB ones.  We used tuning data to optimize these parameters.
\figref{fig:vd-ctr-all} shows how the CTR of each algorithm
changes with respective parameters. All results were obtained by a
single run, but given the size of our dataset and the unbiasedness
result in \thmref{thm:simulation}, the reported numbers are statistically
reliable.

\begin{figure*}[t]
\centering \hspace{-12mm}
\subfigure[\algfont{$\epsilon$-greedy} and
\algfont{ucb}]{\includegraphics[width=0.75\columnwidth]{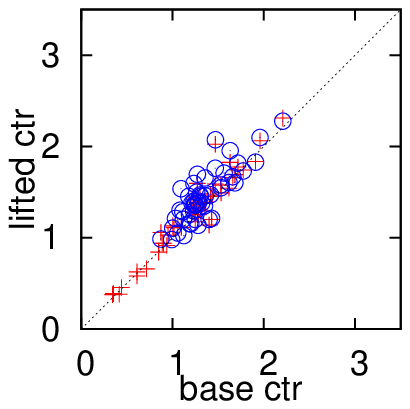}
\label{fig:td-ctrctr-nf}} \hspace{-24mm} \subfigure[seg:\algfont{$\epsilon$-greedy} and
\algfont{ucb}]{\includegraphics[width=0.75\columnwidth]{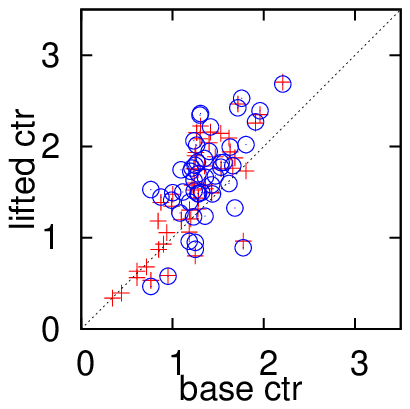}
\label{fig:td-ctrctr-seg}} \hspace{-24mm} \subfigure[disjoint:\algfont{$\epsilon$-greedy} and
\algfont{linucb}]{\includegraphics[width=0.75\columnwidth]{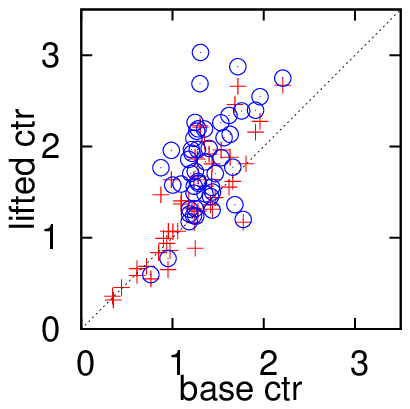}
\label{fig:td-ctrctr-f}} \hspace{-24mm} \subfigure[hybrid:\algfont{$\epsilon$-greedy} and
\algfont{linucb}]{\includegraphics[width=0.75\columnwidth]{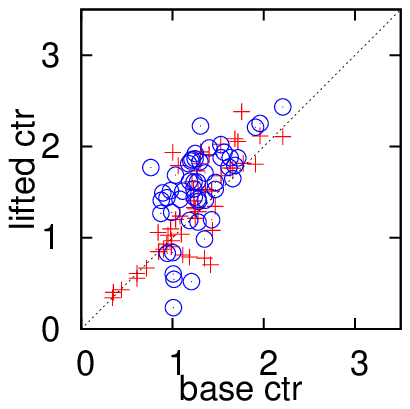}
\label{fig:td-ctrctr-f2}} \hspace{-10mm}
\figsqueeze
\caption{Scatterplots of
the base CTR vs. lifted CTR (in the learning bucket) of the $50$ most frequently selected
articles when $100$\% evaluation data were used.  Red crosses are for $\epsilon$-greedy algorithms, and
blue circles are for UCB algorithms.  Note that the sets of most frequently chosen articles varied with algorithms; see the text for details.} \label{fig:td-ctrctr}
\end{figure*}

First, as seen from \figref{fig:vd-ctr-all}, the CTR curves in the
learning buckets often possess the inverted U-shape.  When the
parameter ($\epsilon$ or $\alpha$) is too small, there was
insufficient exploration, the algorithms failed to identify good
articles, and had a smaller number of clicks.  On the other hand,
when the parameter is too large, the algorithms appeared to
over-explore and thus wasted some of the opportunities to increase
the number of clicks.  Based on these plots on tuning data, we
chose appropriate parameters for each algorithm and ran it once on
the evaluation data in the next subsection.

Second, it can be concluded from the plots that warm-start
information is indeed helpful for finding a better match between
user interest and article content, compared to the no-feature
versions of $\epsilon$-greedy and UCB.  Specifically, both
\algfont{$\epsilon$-greedy (warm)} and \algfont{ucb (warm)} were able
to beat \algfont{omniscient}, the highest CTRs
achievable by context-free policies in hindsight. However,
performance of the two algorithms using warm-start information
is not as stable as algorithms that learn the weights
online. Since the offline model for ``warm start'' was trained
with article CTRs estimated on all random
traffic~\cite{Chu09Personalized}, \algfont{$\epsilon$-greedy
(warm)} gets more stable performance in the deployment bucket when $\epsilon$
is close to $1$. The warm start part also helps \algfont{ucb (warm)}
in the learning bucket by selecting more attractive articles to users from
scratch, but did not help \algfont{ucb (warm)} in determining the
best online for deployment. Since \algfont{ucb} relies on
the a confidence interval for exploration, it is hard to correct the
initialization bias introduced by ``warm start''.
In contrast, all online-learning algorithms
were able to consistently beat the omniscient policy.  Therefore,
we did not try the warm-start algorithms on the evaluation data.

Third, $\epsilon$-greedy algorithms (on the left of
\figref{fig:vd-ctr-all}) achieved similar CTR as upper confidence
bound ones (on the right of \figref{fig:vd-ctr-all}) in the
deployment bucket when appropriate parameters were used.  Thus,
both types of algorithms appeared to learn comparable policies.
However, they seemed to have lower CTR in the learning bucket,
which is consistent with the empirical findings of context-free
algorithms~\cite{Agarwal09Explore} in real bucket tests.

Finally, to compare algorithms when data are sparse, we repeated
the same parameter tuning process for each algorithm with fewer
data, at the level of $30$\%, $20$\%, $10$\%, $5$\%, and $1$\%.
Note that we still used all data to evaluate an algorithm's
CTR as done in \algref{alg:pe}, but then only a fraction of available data were randomly
chosen to be used by the algorithm to improve its policy.

\subsubsection{Results for Evaluation Data} \label{sec:test-results}

With parameters optimized on the tuning data (\cf, \figref{fig:vd-ctr-all}),
we ran the algorithms on the evaluation data and summarized the CTRs in
\tblref{tbl:td-ctr}.  The table also reports the CTR lift compared
to the baseline of \algfont{$\epsilon$-greedy}.  The CTR of \algfont{omniscient}
was $1.615$, and so a significantly larger CTR of an
algorithm indicates its effective use of user/article features for personalization.
Recall that the reported CTRs were normalized by the random policy's
CTR.  We examine the results more closely in the following
subsections.

\subsubsubsection{On the Use of Features}

We first investigate whether it helps to use features in
article recommendation.  It is clear from
\tblref{tbl:td-ctr} that, by considering user features, both
\algfont{$\epsilon$-greedy (seg/disjoint/hybrid)} and
UCB methods (\algfont{ucb (seg)} and \algfont{linucb
(disjoint/hybrid)}) were able to achieve a CTR lift of around
$10$\%, compared to the baseline \algfont{$\epsilon$-greedy}.

To better visualize the effect of features, \figref{fig:td-ctrctr}
shows how an article's CTR (when chosen by an algorithm) was
lifted compared to its base CTR (namely, the context-free
CTR).\footnote{To avoid inaccurate CTR estimates,
only $50$ articles that were chosen most often by
an algorithm were included in its \emph{own} plots.
Hence, the plots for different algorithms are not comparable.}
Here, an article's base CTR measures how interesting it is to a
random user, and was estimated from logged events.  Therefore, a high
ratio of the lifted and base CTRs of an article is a
strong indicator that an algorithm does recommend this article to
potentially interested users. %
\figref{fig:td-ctrctr-nf} shows neither
\algfont{$\epsilon$-greedy} nor \algfont{ucb} was able to lift
article CTRs, since they made no use of user information.  In
contrast, all the other three plots show clear benefits by
considering personalized recommendation.  In an extreme case
(\figref{fig:td-ctrctr-f}), one of the article's CTR was lifted
from $1.31$ to $3.03$---a $132$\% improvement.%

Furthermore, it is consistent with our previous results on
tuning data that, compared to $\epsilon$-greedy algorithms,
UCB methods achieved higher CTRs in the
deployment bucket, and the advantage was even greater in the
learning bucket.
As mentioned in \secref{sec:existing-algorithms},
$\epsilon$-greedy approaches are \emph{unguided} because they
choose articles \emph{uniformly} at random for exploration.  In
contrast, exploration in upper confidence bound methods are
effectively \emph{guided} by confidence intervals---a measure of
uncertainty in an algorithm's CTR estimate.  Our experimental
results imply the effectiveness of upper confidence bound methods
and we believe they have similar benefits in many other
applications as well.

\subsubsubsection{On the Size of Data}

One of the challenges in personalized web services is the scale of
the applications.  In our problem, for example, a small pool of news
articles were hand-picked by human editors.  But if we wish to
allow more choices or use automated article selection methods to
determine the article pool, the number of articles can be too
large even for the high volume of Yahoo!\ traffic.  Therefore, it
becomes critical for an algorithm to quickly identify a good match
between user interests and article contents when data are sparse.
In our experiments, we artificially reduced data size (to the
levels of $30$\%, $20$\%, $10$\%, $5$\%, and $1$\%, respectively)
to mimic the situation where we have a large article pool but a
fixed volume of traffic.

\begin{figure}[t]
\centering \subfigure[CTRs in the deployment
bucket.]{\includegraphics[angle=0,width=\columnwidth]{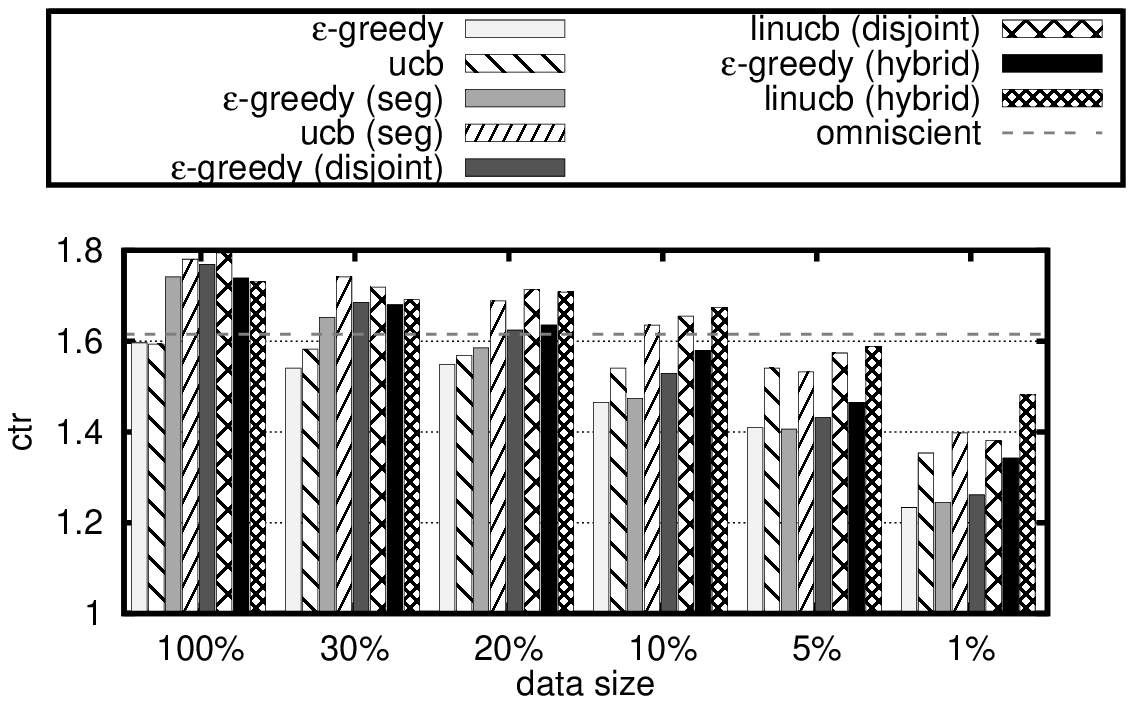}
\label{fig:td-expctr-s}} \subfigure[CTRs in the learning
bucket.]{\includegraphics[angle=0,width=\columnwidth]{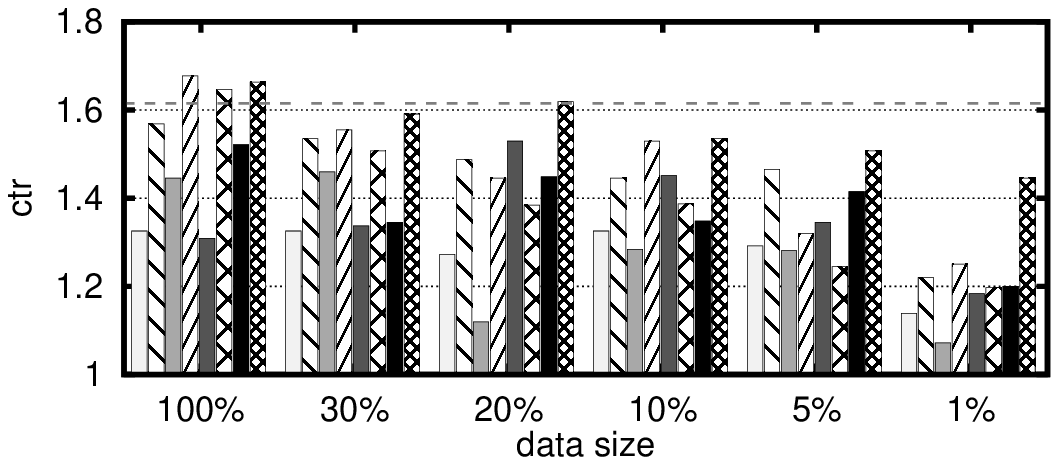}
\label{fig:td-ovrctr-s}}
\figsqueeze
\caption{CTRs in evaluation data with varying
data sizes.} \label{fig:td-ctr-s}
\end{figure}

To better visualize the comparison results, we use bar graphs in
\figref{fig:td-ctr-s} to plot all algorithms' CTRs with various
data sparsity levels.  A few observations are in order.  First, at
\emph{all} data sparsity levels, features were still useful.  At
the level of $1$\%, for instance, we observed a $10.3$\%
improvement of \algfont{linucb (hybrid)}'s CTR in the deployment
bucket ($1.493$) over \algfont{ucb}'s ($1.354$).

Second, UCB methods consistently outperformed
$\epsilon$-greedy ones in the deployment bucket.\footnote{In the
less important learning bucket, there were two exceptions for
\algfont{linucb (disjoint)}.}  The advantage over
$\epsilon$-greedy was even more apparent when data size was
smaller.

Third, compared to \algfont{ucb (seg)} and \algfont{linucb
(disjoint)}, \algfont{linucb (hybrid)} showed significant benefits
when data size was small.  Recall that in hybrid models, some features
are shared by all articles, making it
possible for CTR information of one article to be ``transferred''
to others.  This advantage is particularly useful when the
article pool is large.  In contrast, in disjoint models, feedback of
one article may not be utilized by other articles; the same is
true for \algfont{ucb (seg)}.  \figref{fig:td-expctr-s} shows
transfer learning is indeed helpful when data are sparse.

\subsubsubsection{Comparing \algfont{ucb (seg)} and \algfont{linucb (disjoint)}}

From \figref{fig:td-expctr-s}, it can be seen that \algfont{ucb
(seg)} and \algfont{linucb (disjoint)} had similar
performance.  We believe it was no coincidence.  Recall that
features in our disjoint model are actually normalized membership measures of a
user in the five clusters described in \secref{sec:features}.
Hence, these features may be viewed as a ``soft'' version of
the user assignment process adopted by \algfont{ucb (seg)}.

\begin{figure}[t]
\centering
\includegraphics[width=0.9\columnwidth]{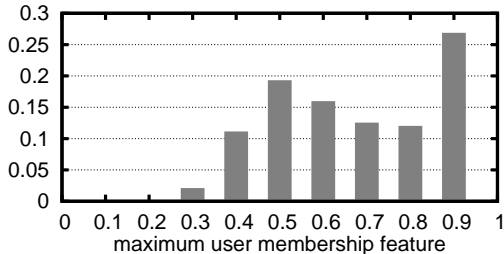}
\figsqueeze
\caption{User maximum membership histogram.} %
\label{fig:td-user-hist}
\end{figure}

\figref{fig:td-user-hist} plots the histogram of a user's relative
membership measure to the closest cluster, namely, the largest
component of the user's five, non-constant features.  It is clear
that most users were quite close to one of the five cluster
centers: the maximum membership of about $85$\% users were higher
than $0.5$, and about $40$\% of them were higher than $0.8$.
Therefore, many of these features have a highly dominating
component, making the feature vector similar to the ``hard'' version
of user group assignment.

We believe that adding more features with diverse components, such
as those found by principal component analysis, would be necessary
to further distinguish \algfont{linucb (disjoint)} from \algfont{ucb
(seg)}. \vspace{-1mm}

\comment{We believe that adding more features from various facets
should be helpful in distinguishing both ``borderline'' users and
``hardcore'' users.}

\section{Conclusions}

This paper takes a contextual-bandit approach to personalized web-based services
such as news article recommendation.  We proposed a simple and reliable
method for evaluating bandit algorithms directly from logged events,
so that the often problematic simulator-building step could be avoided.
Based on real Yahoo!\ Front Page traffic, we found that upper confidence
bound methods generally outperform the simpler yet unguided $\epsilon$-greedy
methods.  Furthermore, our new algorithm \algfont{LinUCB} shows advantages
when data are sparse, suggesting its effectiveness to personalized
web services when the number of contents in the pool is large.

In the future, we plan to investigate bandit approaches to other
similar web-based serviced such as online advertising, and compare our algorithms
to related methods such as Banditron~\cite{Kakade08Efficient}.
A second direction
is to extend the bandit formulation and algorithms in which an ``arm''
may refer to a complex object rather than an item (like an article).
An example is ranking, where an arm corresponds to a permutation of
retrieved webpages.  Finally, user interests change over time, and so
it is interesting to consider temporal information in bandit algorithms.

\section{Acknowledgments}

We thank Deepak Agarwal, Bee-Chung Chen, Daniel Hsu, and Kishore Papineni for many helpful discussions, Istv\'{a}n Szita and Tom Walsh for clarifying their algorithm, and Taylor Xi and the anonymous reviewers for suggestions that improved the presentation of the paper.

{\small\bibliographystyle{abbrv}
\bibliography{refs}}  %

\end{document}

%% file: math.tex
\newcommand{\E}{{\operatorname{\mathbf{E}}}}  

\newcommand{\mi}{{-1}}  
\newcommand{\mt}{{\top}}  

\newcommand{\vecb}[1]{{\mathbf{#1}}}
\newcommand{\abs}[1]{\left\lvert#1\right\rvert}  
\newcommand{\Rset}{{\mathbb{R}}}  

\newcommand{\defeq}{{\stackrel{\mathrm{def}}{=}}}

%% file: td-ctr-all.tex
\begin{tabular}{|c|c|c|c|c|c|c|c|c|c|c|c|c|}
\hline
\multirow{2}{*}{algorithm} & \multicolumn{2}{|c|}{size = 100\%} & \multicolumn{2}{|c|}{size = 30\%} & \multicolumn{2}{|c|}{size = 20\%} & \multicolumn{2}{|c|}{size = 10\%} & \multicolumn{2}{|c|}{size = 5\%} & \multicolumn{2}{|c|}{size = 1\%} \\
\cline{2-13} & deploy & learn & deploy & learn & deploy & learn & deploy & learn & deploy & learn & deploy & learn \\
\hline\hline
\multirow{2}{*}{\algfont{$\epsilon$-greedy}} & $1.596$ & $1.326$ & $1.541$ & $1.326$ & $1.549$ & $1.273$ & $1.465$ & $1.326$ & $1.409$ & $1.292$ & $1.234$ & $1.139$\\
 & $0$\% & $0$\% & $0$\% & $0$\% & $0$\% & $0$\% & $0$\% & $0$\% & $0$\% & $0$\% & $0$\% & $0$\% \\
\hline
\multirow{2}{*}{\algfont{ucb}} & $1.594$ & $1.569$ & $1.582$ & $1.535$ & $1.569$ & $1.488$ & $1.541$ & $1.446$ & $1.541$ & $1.465$ & $1.354$ & $1.22$\\
 & $0$\% & $18.3$\% & $2.7$\% & $15.8$\% & $1.3$\% & $16.9$\% & $5.2$\% & $9$\% & $9.4$\% & $13.4$\% & $9.7$\% & $7.1$\% \\
\hline
\multirow{2}{*}{\algfont{$\epsilon$-greedy (seg)}} & $1.742$ & $1.446$ & $1.652$ & $1.46$ & $1.585$ & $1.119$ & $1.474$ & $1.284$ & $1.407$ & $1.281$ & $1.245$ & $1.072$\\
 & $9.1$\% & $9$\% & $7.2$\% & $10.1$\% & $2.3$\% & $-12$\% & $0.6$\% & $-3.1$\% & $0$\% & $-0.8$\% & $0.9$\% & $-5.8$\% \\
\hline
\multirow{2}{*}{\algfont{ucb (seg)}} & $1.781$ & $1.677$ & $1.742$ & $1.555$ & $1.689$ & $1.446$ & $1.636$ & $1.529$ & $1.532$ & $1.32$ & $1.398$ & $1.25$\\
 & $11.6$\% & $26.5$\% & $13$\% & $17.3$\% & $9$\% & $13.6$\% & $11.7$\% & $15.3$\% & $8.7$\% & $2.2$\% & $13.3$\% & $9.7$\% \\
\hline
\multirow{2}{*}{\algfont{$\epsilon$-greedy (disjoint)}} & $1.769$ & $1.309$ & $1.686$ & $1.337$ & $1.624$ & $1.529$ & $1.529$ & $1.451$ & $1.432$ & $1.345$ & $1.262$ & $1.183$\\
 & $10.8$\% & $-1.2$\% & $9.4$\% & $0.8$\% & $4.8$\% & $20.1$\% & $4.4$\% & $9.4$\% & $1.6$\% & $4.1$\% & $2.3$\% & $3.9$\% \\
\hline
\multirow{2}{*}{\algfont{linucb (disjoint)}} & $1.795$ & $1.647$ & $1.719$ & $1.507$ & $1.714$ & $1.384$ & $1.655$ & $1.387$ & $1.574$ & $1.245$ & $1.382$ & $1.197$\\
 & $12.5$\% & $24.2$\% & $11.6$\% & $13.7$\% & $10.7$\% & $8.7$\% & $13$\% & $4.6$\% & $11.7$\% & $-3.5$\% & $12$\% & $5.1$\% \\
\hline
\multirow{2}{*}{\algfont{$\epsilon$-greedy (hybrid)}} & $1.739$ & $1.521$ & $1.68$ & $1.345$ & $1.636$ & $1.449$ & $1.58$ & $1.348$ & $1.465$ & $1.415$ & $1.342$ & $1.2$\\
 & $9$\% & $14.7$\% & $9$\% & $1.4$\% & $5.6$\% & $13.8$\% & $7.8$\% & $1.7$\% & $4$\% & $9.5$\% & $8.8$\% & $5.4$\% \\
\hline
\multirow{2}{*}{\algfont{linucb (hybrid)}} & $1.73$ & $1.663$ & $1.691$ & $1.591$ & $1.708$ & $1.619$ & $1.675$ & $1.535$ & $1.588$ & $1.507$ & $1.482$ & $1.446$\\
 & $8.4$\% & $25.4$\% & $9.7$\% & $20$\% & $10.3$\% & $27.2$\% & $14.3$\% & $15.8$\% & $12.7$\% & $16.6$\% & $20.1$\% & $27$\% \\
\hline
\end{tabular}